\pdfoutput=1

\documentclass[11pt]{article}

\usepackage{ACL2023}

\usepackage{times}
\usepackage{latexsym}
\usepackage[T1]{fontenc}
\usepackage[utf8]{inputenc}
\usepackage{microtype}
\usepackage{inconsolata}

\usepackage[utf8]{inputenc}
\usepackage[pdftex]{graphicx}
\usepackage{booktabs}
\usepackage{amsmath}
\usepackage{amssymb}
\usepackage{float}
\usepackage{caption}
\usepackage{subcaption}
\usepackage{multirow}

\title{Beyond Turing: A Comparative Analysis of Approaches for Detecting Machine-Generated Text}

\author{{Muhammad Farid Adilazuarda}\thanks{*Work conducted while visiting University of Zagreb.} \\
  MBZUAI \quad Institut Teknologi Bandung  \\
   University of Zagreb, Faculty of Electrical Engineering and Computing\\
  \texttt{farid.adilazuarda@mbzuai.ac.ae}}

\begin{document}
\maketitle
\begin{abstract}
Significant progress has been made on text generation by pre-trained language models (PLMs), yet distinguishing between human and machine-generated text poses an escalating challenge. This paper offers an in-depth evaluation of three distinct methods used to address this task: traditional shallow learning, Language Model (LM) fine-tuning, and Multilingual Model fine-tuning. These approaches are rigorously tested on a wide range of machine-generated texts, providing a benchmark of their competence in distinguishing between human-authored and machine-authored linguistic constructs. The results reveal considerable differences in performance across methods, thus emphasizing the continued need for advancement in this crucial area of NLP. This study offers valuable insights and paves the way for future research aimed at creating robust and highly discriminative models.
\end{abstract}

\section{Introduction}
The drive to discern between human and machine-generated text has been a long-standing pursuit, tracing its origins back to Turing's famous 'Turing Test', which explore a machine's ability to imitate human-like intelligence. With the vast and rapid development of advanced PLMs, the capacity to generate increasingly human-like text has grown, blurring the lines of detectability and bringing this research back into sharp focus.

Addressing this complexity, this paper explores two specific tasks: 1) the differentiation between human and machine-generated text, and 2) the identification of the specific language model that generated a given text. Our exploration extends beyond the traditional shallow learning techniques, exploring into the more robust methodologies of Language Model (LM) fine-tuning and Multilingual Model fine-tuning \citep{winata2021language, adilazuarda2023indorobusta, radford2019language}. These techniques enable PLMs to specialize in the detection and categorization of machine-generated texts. They adapt pre-existing knowledge to the task at hand, effectively manage language-specific biases, and improve classification performance. Note that in this experiment, we do not use parameter-efficient strategies even when they have a superior specific-language capabilities. This is due to our constraint to fully fine-tune a language model and given the modular models' limited capabilities in such tasks \cite{adilazuarda2023obscure}.

Through an exhaustive examination of a diverse set of machine-generated texts, Our paper offers the following contributions:
\begin{enumerate}
    \itemsep0em 
    \item An exhaustive evaluation of the capabilities of PLMs in categorizing machine-generated texts.
    \item An investigation into the effectiveness of employing multilingual techniques to mitigate language-specific biases in the detection of machine-generated text.
    \item The application of a few-shot multilingual evaluation strategy to measure the adaptability of models in resource-limited scenarios.
\end{enumerate}

\section{Related Works}

This study's related work falls into three main categories: machine-generated text detection, identification of specific PLMs, and advancements in language model fine-tuning.

\textbf{Machine-generated Text Detection:} Distinguishing human from machine-generated text has become an intricate challenge with recent advancements in language modeling. Prior research \citep{schwartz2018effect,ippolito2020discriminating, jawahar2020automatic, he2024mgtbench, tian2023multiscale, bhattacharjee2023fighting, koike2023outfox, yu2023gpt} has explored nuances separating human and machine compositions. Our work builds on these explorations by assessing various methodologies for this task.

\textbf{Language Models Identification:} Some studies \citep{antoun2023robust, guo2023close, wu2023llmdet, mitchell2023detectgpt, deng2023efficient, su2023detectllm, li2023deepfake, liu2023argugpt, chen2023gptsentinel} attempt to identify the specific language model generating a text. These efforts, however, are still in growing stages and often rely on model-specific features. Our work evaluates various methods' efficacy for this task, focusing on robustness across a spectrum of PLMs.

\textbf{Language Model Fine-tuning Advances:} Language Model fine-tuning \citep{howard2018universal} and Multilingual Model fine-tuning \citep{conneau2020massively} represent progress in language model customization. They enable model specialization in machine-generated text detection and classification and address language-specific biases, thereby enhancing classification accuracy across diverse languages.

This study intertwines these three research avenues, providing a thorough evaluation of the mentioned methodologies in machine-generated text detection and classification.

\subsection{Dataset}
Our experiments utilize two multi-class classification datasets, namely Subtask 1 and Subtask 2, as referenced from the publicly available Autextification dataset~\cite{autextification}. Subtask 1 is a document-level dataset composed of \textbf{65,907} samples. Each sample is assigned one of two class labels: 'generated' or 'human'. Subtask 2, serves as a Model Attribution dataset consisting of \textbf{44,351} samples. This dataset includes six different labels - A, B, C, D, E, and F - representing distinct models of text generation. A detailed overview of the statistics related to both Subtask 1 and Subtask 2 datasets is provided in Table~\ref{tab:dataset_statistics}.

\begin{table}[ht]
\centering
\resizebox{\columnwidth}{!}{
\begin{tabular}{llcccc}
\toprule
\textbf{Language} & \textbf{Subtask} & $|$\textbf{Train}$|$ & $|$\textbf{Valid}$|$ & $|$\textbf{Test}$|$  & \textbf{\#Class} \\ \midrule
\multirow{2}{*}{\textbf{English}} & Subtask 1 & 27,414 & 3,046 & 3,385 & 2 \\
 & Subtask 2 & 18,156 & 2,018 & 2,242 & 6 \\ 
\midrule
\multirow{2}{*}{\textbf{Spanish}} & Subtask 1 & 25,969 & 2,886 & 3,207 & 2 \\
 & Subtask 2 & 17,766 & 1,975 & 2,194 & 6 \\ 
\bottomrule
\end{tabular}
}
\caption{Statistics of the datasets.}
\label{tab:dataset_statistics}
\end{table}

\section{Methods}
\subsection{Shallow Learning}
We conducted an evaluation of two distinct shallow learning models, specifically Logistic Regression and XGBoost, utilizing Fasttext word embeddings that were trained on our preprocessed training set. FastText's subword representation captures fine morphological details. This is useful in detecting differences between the often overly formal structured machine-generated text and the morphologically rich human-generated text. 

Prior to the training process, we implemented a fundamental preprocessing step involving non-ASCII and special characters removal. As showed in Table \ref{tab:feature_engineering}, we propose embedding on four lexical complexity measures aimed at quantifying different aspects of a text:

\textbf{Average Word Length (AWL)}: This metric reflects the lexical sophistication of a text, with longer average word lengths potentially suggesting more complex language use. Let $W = \{w_1, w_2, ..., w_n\}$ represent the set of word tokens in the text. The $AWL$ is given by:
\[
AWL = \frac{1}{n} \sum_{i=1}^{n} |w_i|
\]

\textbf{Average Sentence Length (ASL)}: This measures syntactic complexity, with longer sentences often requiring more complex syntactic structures.Let $S = \{s_1, s_2, ..., s_m\}$ represent the set of sentence tokens in the text. The $ASL$ is defined as:
\[
ASL = \frac{1}{m} \sum_{j=1}^{m} |s_j|
\]

\textbf{Vocabulary Richness (VR)}: This ratio of unique words to the total number of words is a measure of lexical diversity. If $UW$ represents the set of unique words in the text, the $VR$ is calculated as:
\[
VR = \frac{|UW|}{n}
\]

\textbf{Repetition Rate (RR)}: The ratio of words occurring more than once to the total number of words, indicative of the redundancy of a text. If $RW$ represents the set of words that occur more than once, $RR$ is computed as:
\[
RR = \frac{|RW|}{n}
\]

Table \ref{tab:feature_engineering} presents a snapshot of our dataset after the application of our feature calculations. These include Average Word Length \textbf{(AWL)}, Average Sentence Length \textbf{(ASL)}, Vocabulary Richness \textbf{(VR)}, and Repetition Rate \textbf{(RR)}. By computing these features, we aimed to capture distinct textual characteristics that could aid our models in discriminating human and machine-generated text.

\begin{table}[ht]
\centering
\resizebox{\columnwidth}{!}{%
\begin{tabular}{p{1.5cm}cccccc}
\toprule
\textbf{Text} & \textbf{Label} & \textbf{AWL} & \textbf{ASL} & \textbf{VR} & \textbf{RR} \\
\midrule
\text{you need to...} & generated & 3.12 & 49.50 & 0.96 & 0.04 \\
\text{The Comm...} & generated & 4.92 & 62.56 & 0.69 & 0.09 \\
\text{I pass my...} & human & 3.55 & 90.00 & 0.90 & 0.10 \\
\bottomrule
\end{tabular}
}
\caption{Text feature calculation. Label, AWL: Avg. Word Length, ASL: Avg. Sent. Length, VR: Vocab. Richness, RR: Repetition Rate}
\label{tab:feature_engineering}
\end{table}

\subsection{Language Model Finetuning}
In this study, we employed multiple models: XLM-RoBERTa, mBERT, DeBERTa-v3, BERT-tiny, DistilBERT, RoBERTa-\textit{Detector}, and ChatGPT-\textit{Detector}. The models were fine-tuned on single and both languages simultaneously using multilingual training \citep{bai2021joint}.

During evaluation, we employed the F1 score for our primary metrics. Furthermore, we incorporated a Few-Shot learning evaluation to assess our models' capacity to learn effectively from a limited set of examples for their practical applicability in real-world scenarios. This involved using varying seed quantities of \texttt{[200, 400, 600, 800, 1000]} instances, applied across both English and Spanish languages.

\section{Experiments}
Our approach to fine-tuning PLMs remained consistent across all models under consideration. We utilized HuggingFace's Transformers library\footnote{https://huggingface.co/}, which provides both pre-trained models and scripts for fine-tuning. Utilizing a multi-GPU setup, we employed the AdamW optimizer \citep{loshchilov2019decoupled}, configured with a learning rate of \texttt{1e-6} and a batch size of 64. To prevent overfitting, we implemented early stopping within 3 epochs patience. The models were trained across a total of 10 epochs.

\textbf{Multilingual Finetuning}. An integral part of our approach was the models fine-tuning using English and Spanish data to capture the unique linguistic features of each language.

\begin{figure}[!ht]
     \centering
     \begin{subfigure}[b]{0.23\textwidth}
         \centering
         \includegraphics[width=\textwidth]{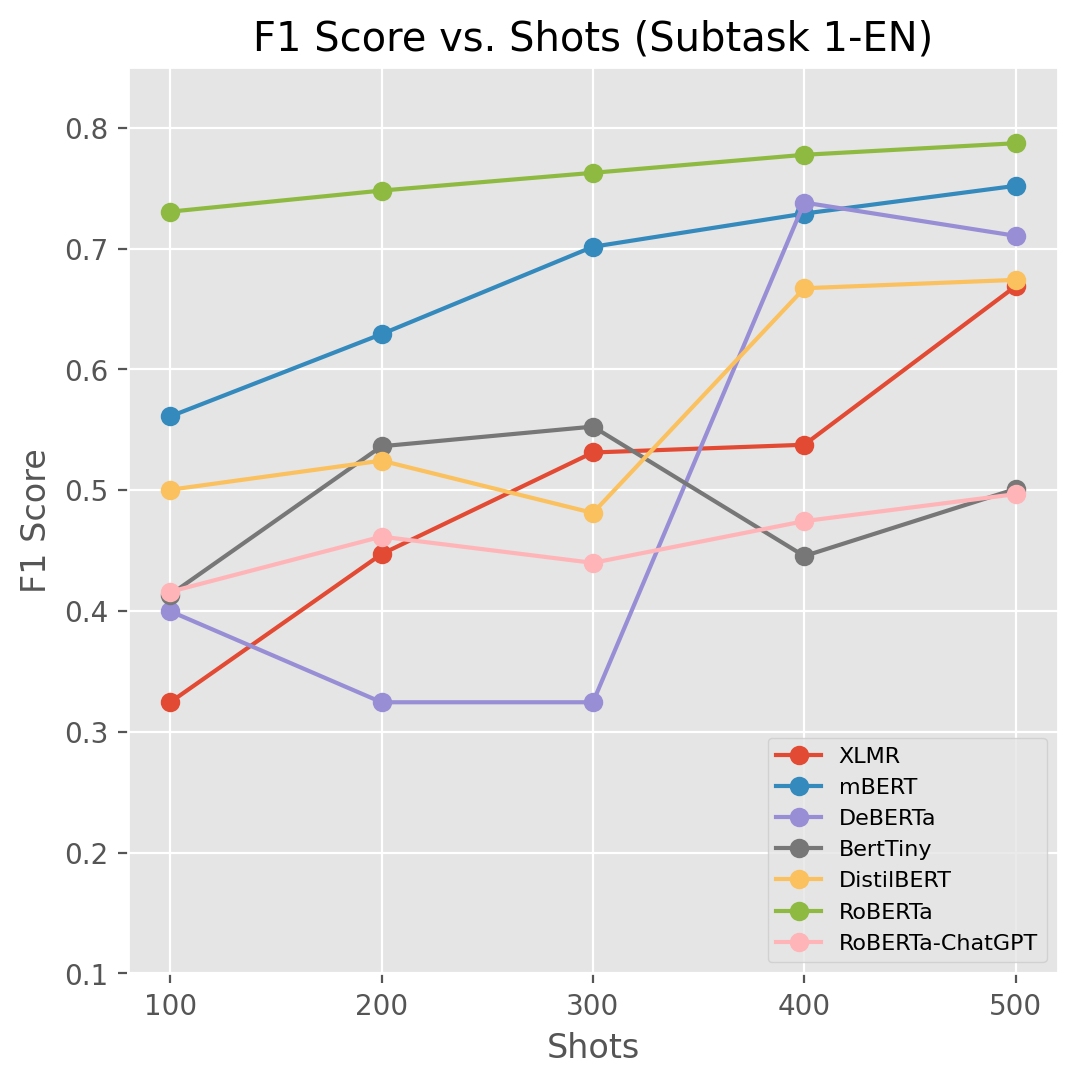}
         \caption{English}
         \label{fig:fewshot_eval_subtask1_en}
     \end{subfigure}
     \begin{subfigure}[b]{0.23\textwidth}
         \centering
         \includegraphics[width=\textwidth]{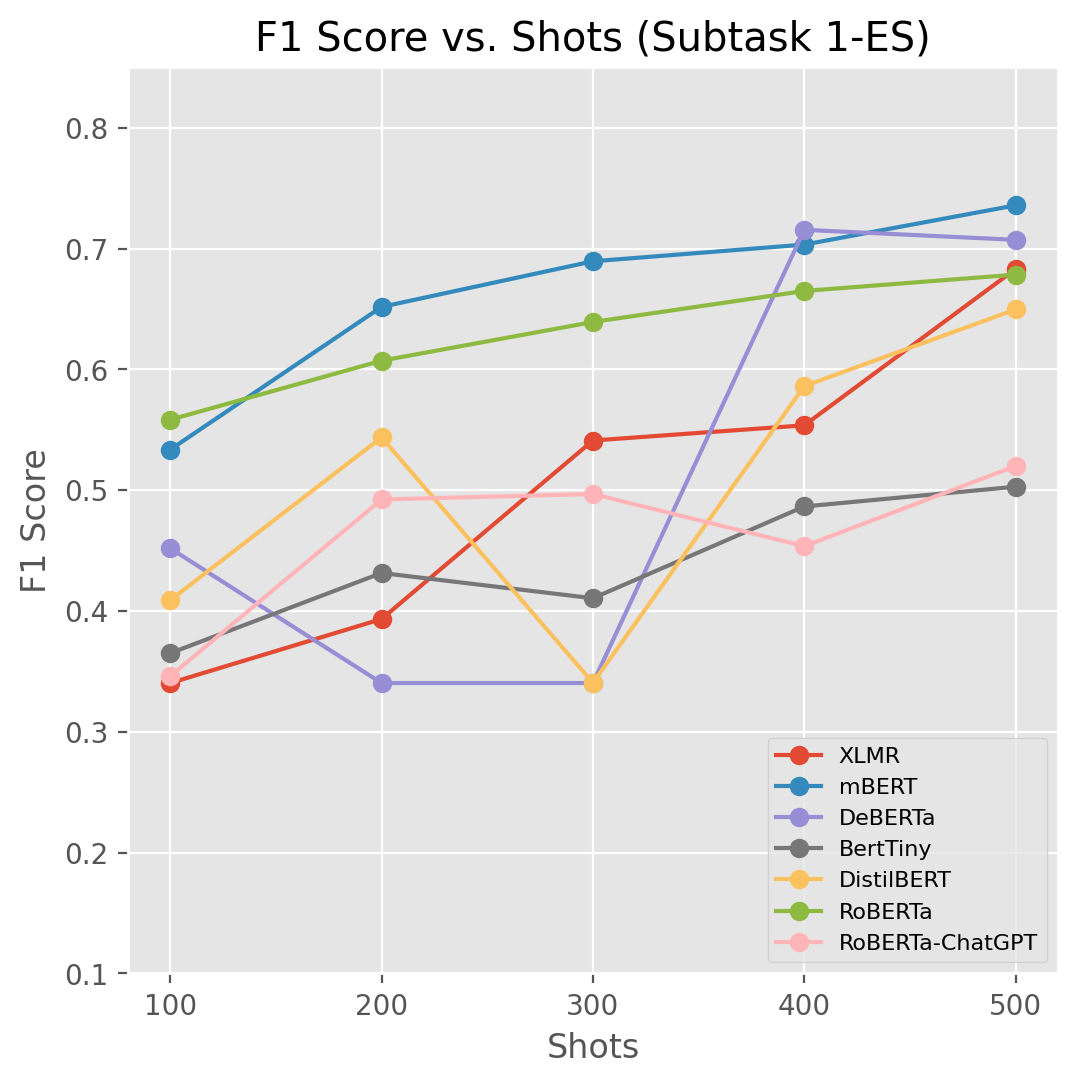}
         \caption{Spanish}
         \label{fig:fewshot_eval_subtask1_es}
     \end{subfigure}
        \caption{Subtask 1 Evaluation on Few-Shot Learning}
        \label{fig:few_shot_subtask1}
\end{figure}

\textbf{Few-Shot Learning}. To see the performance of the models in few-shot learning scenarios, employ few-shot learning experiments ranging from 200 to 1000 samples combination from the English and Spanish training data. The results of the few-shot learning experiments are depicted in Fig. \ref{fig:few_shot_subtask1}.


\section{Results and Discussion}
\subsection{Distinguishing Capability}
From the few-shot learning experiments, the models' performance varied significantly in distinguishing between human and machine-generated text.  In the default evaluation, multilingually-finetuned mBERT outperformed the other models in English, and single-language finetuned mBERT exhibited the highest score in Spanish. However, In the few-shot experiment setting, the RoBERTa-\textit{Detector} demonstrated the most robust distinguishing capability, scoring up to 0.787 with 1000 samples.

When comparing these results, we can observe that mBERT maintains strong performance in both the few-shot learning experiments and the single language experiments. It suggests that mBERT could provide a reliable choice across different tasks and experimental settings in both Subtasks.

\subsection{Model Generation Capability}

\begin{table}[!ht]
\centering
\resizebox{1\linewidth}{!}{
\begin{tabular}{lcccccc}
\toprule
\textbf{Model} & A & B & C & D & E & F \\
\midrule
\textbf{Error(\%)} & 37.62 & 68.43 & 58.55 & 48.89 & \textbf{74.24} & 13.81 \\
\bottomrule
\end{tabular}
}
\caption{Comparison of Model Error Percentages. The models, labeled as A, B, C, D, E, and F, were used for prediction. The error rate was computed using mBERT with multilingual fine-tuning.}
\label{tab:error_percentage}
\end{table}

\begin{table*}[!ht]
\centering
\resizebox{0.6\linewidth}{!}{
\begin{tabular}{l|cc|cc}
\toprule
\multicolumn{1}{c|}{\bf Model} & \multicolumn{2}{c|}{\bf Subtask 1} & \multicolumn{2}{c}{\bf Subtask 2} \\
 & \textbf{English-F1} & \textbf{Spanish-F1} & \textbf{English-F1} & \textbf{Spanish-F1} \\
\midrule
\multicolumn{5}{c}{\textit{\textbf{Shallow Learning + Feat. Engineering}}} \\
\midrule
Logistic Regression & 65.67\% & 63.87\% & 38.39\% & 42.99\% \\
XGBoost & 71.52\% & 71.53\% & 38.47\% & 41.08\% \\
\midrule
\multicolumn{5}{c}{\textit{\textbf{Fine-tuning}}} \\
\midrule
XLM-RoBERTa & 78.80\% & 76.56\% & 27.14\% & 30.66\% \\
mBERT & \underline{85.18}\% & \textbf{83.25}\% & \underline{44.82}\% & \underline{45.16}\% \\
DeBERTa-V3 & 81.52\% & 72.58\% & 43.93\% & 28.28\% \\
TinyBERT & 63.75\% & 57.83\% & 15.38\% & 13.02\% \\
DistilBERT & 84.97\% & 78.77\% & 41.53\% & 35.61\% \\
RoBERTa-\textit{Detector} & 84.01\% & 75.18\% & 34.13\% & 22.10\% \\
ChatGPT-\textit{Detector} & 68.33\% & 64.64\% & 23.84\% & 25.45\% \\
\midrule
\multicolumn{5}{c}{\textit{\textbf{Multilingual Finetuning}}} \\
\midrule
mBERT & 84.80\% & \underline{82.99}\% & \textbf{49.24}\% & \textbf{47.28}\% \\
DistilBERT & \textbf{85.22}\% & 80.49\% & 41.64\% & 35.59\% \\
\bottomrule
\end{tabular}
}
\caption{F1 Score for Various Models in English and Spanish for Subtask 1 and 2. \textbf{Bold} and \underline{underline} denote first and second best, respectively.}
\label{tab:english_spanish_f1_subtasks}
\end{table*}

Figure \ref{tab:error_percentage} illustrates the error rates of the evaluated models, with \textbf{Model E} has the highest error rate at 74.24\%. In this context, a higher error rate is interpreted positively, indicating that Model E has the strongest capability to generate deceptive text. This could mean that Model E is best at creating text that is complex or nuanced enough to trick the detector into making incorrect judgments. \textbf{Model F}, conversely, shows the lowest error rate at 13.81\%. This suggests that it is the least capable at generating deceptive text compared to the other models. It might produce more predictable or simpler text that the detector can easily identify as generated, hence fewer errors in detection.

However, it's worth noting that the performance might be influenced by "similarity bias in architecture" between the detector and generator models. This means if the generator and detector models are structurally similar, they might share certain biases or weaknesses, which could skew the error rates. For instance, if both models are based on a similar underlying technology (like a specific version of BERT adapted for multilingual contexts, mentioned as mBERT with multilingual fine-tuning), they might inherently perform similarly in certain tasks or languages, affecting the observed error rates.

\subsection{Comparative Analysis of Model Performances}
Our analysis from experiments in Table \ref{tab:english_spanish_f1_subtasks} reveals variations in the performance of the models for both tasks: differentiating human and machine-generated text, and identifying the specific language model that generated the given text. For the first task, mBERT emerges as the top performer with English and Spanish F1 scores of 85.18\% and 83.25\% respectively, in the fine-tuning setup. This performance is closely followed by DistilBERT's English F1 score of 84.97\% and Spanish score of 78.77\%. In the multilingual fine-tuning configuration, DistilBERT edges out with an English F1 score of 85.22\%, but mBERT retains its high Spanish performance with an F1 score of 82.99\%.

In the second task, mBERT continues to excel, achieving F1 scores of 44.82\% and 45.16\% for English and Spanish respectively in the fine-tuning setup. It improves further in the multilingual fine-tuning setup with English and Spanish scores of 49.24\% and 47.28\%. However, models such as XLM-RoBERTa and TinyBERT show substantial performance gaps between the tasks. For example, XLM-RoBERTa excels in the first task with English and Spanish F1 scores of 78.8\% and 76.56\%, but struggles with the second task, with F1 scores dropping to 27.14\% and 30.66\%. Similarly, TinyBERT shows a notable performance drop in the second task.

The performance disparity suggests that the two tasks require distinct skills: the first relies on detecting patterns unique to machine-generated text, while the second demands recognition of nuanced characteristics of specific models. In conclusion, mBERT demonstrates a consistent and robust performance across both tasks. However, the findings also underscore a need for specialized models or strategies for each task, paving the way for future work in the design and fine-tuning of models for these tasks.



\section{Conclusion}
This study performed an exhaustive investigation into three distinct methodologies: traditional shallow learning, Language Model fine-tuning, and Multilingual Model fine-tuning, for detecting machine-generated text and identifying the specific language model that generated the text. Our findings showed that mBERT is a robust discriminator model across different tasks and settings. However, other models like XLM-RoBERTa and TinyBERT showed a noticeable performance gap between the tasks, indicating that these two tasks might require different skillsets. This research provides insights into the performance of these methodologies on a diverse set of machine-generated texts. It also highlights the critical importance of developing specialized models or strategies for each task.


\section*{Limitations}
This study provides a comprehensive comparison and analysis of models' abilities to distinguish between human and machine-generated texts. However, it relies on datasets from the Autextification competition, which withholds the specific models used for text generation in Subtask 1. As a result, in Subtask 2, our classification is based on anonymous labels (A, B, C, D, E, F), without insight into the actual models. This lack of transparency limits our assessment of potential data biases or architectural effects on the classification results. Future work that overcomes these limitations could enhance the depth and accuracy of the analysis.

\section*{Acknowledgements}

We express our profound gratitude to our mentors, Professor Jan Šnajder and Teaching Assistant Josip Jukić, for their invaluable guidance, constructive feedback, and unwavering support throughout the duration of this project. Their expertise and dedication have significantly contributed to the advancement of our research and understanding.


\bibliography{anthology,custom}
\bibliographystyle{acl_natbib}

\appendix
\clearpage
\section{Dataset Statistics}
\begin{figure*}[!h]
     \centering
     \begin{subfigure}[b]{0.45\textwidth}
         \centering
         \includegraphics[width=\textwidth]{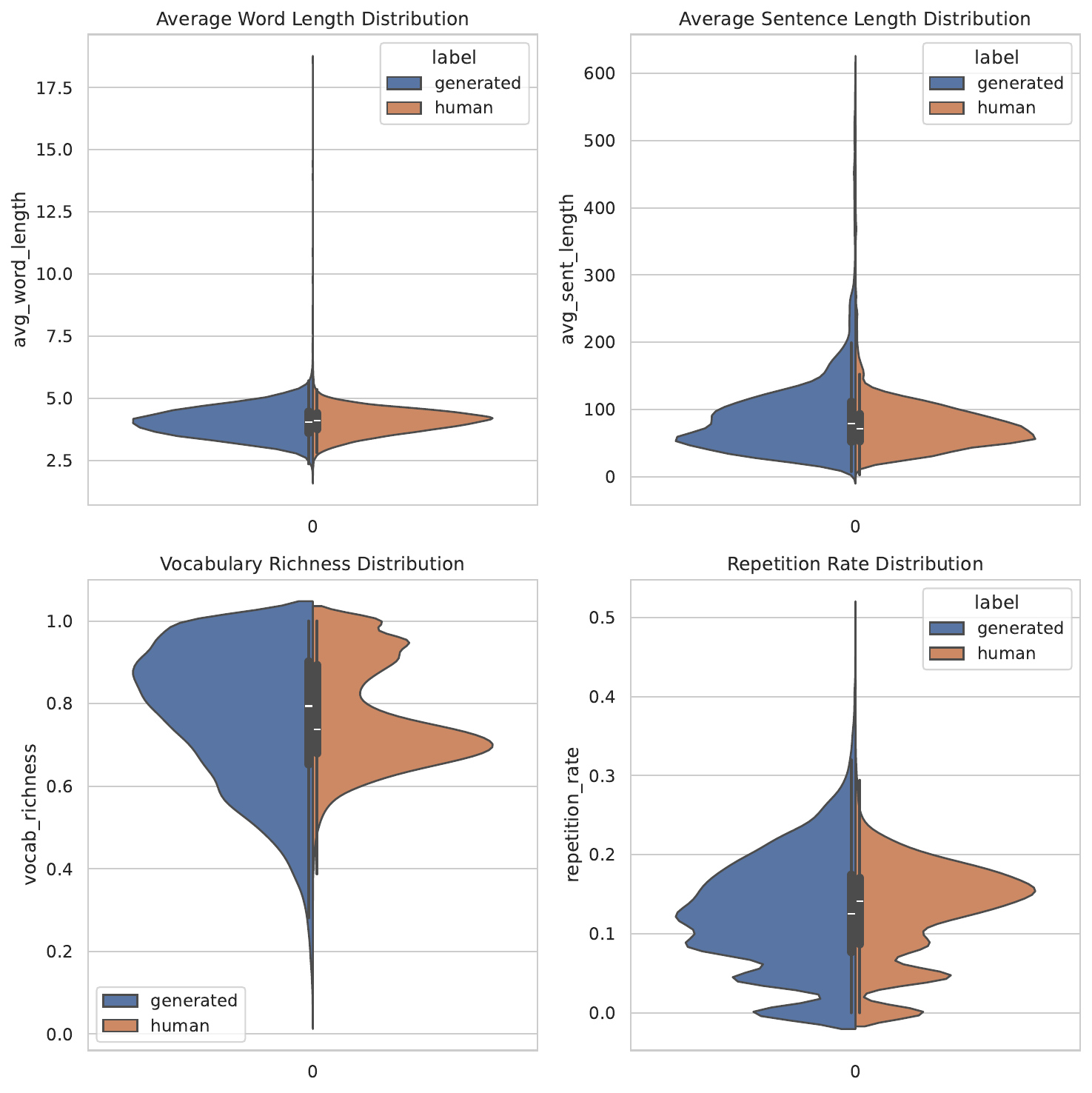}
         \caption{English}
         \label{fig:EDA_subtask1_en}
     \end{subfigure}
     \begin{subfigure}[b]{0.45\textwidth}
         \centering
         \includegraphics[width=\textwidth]{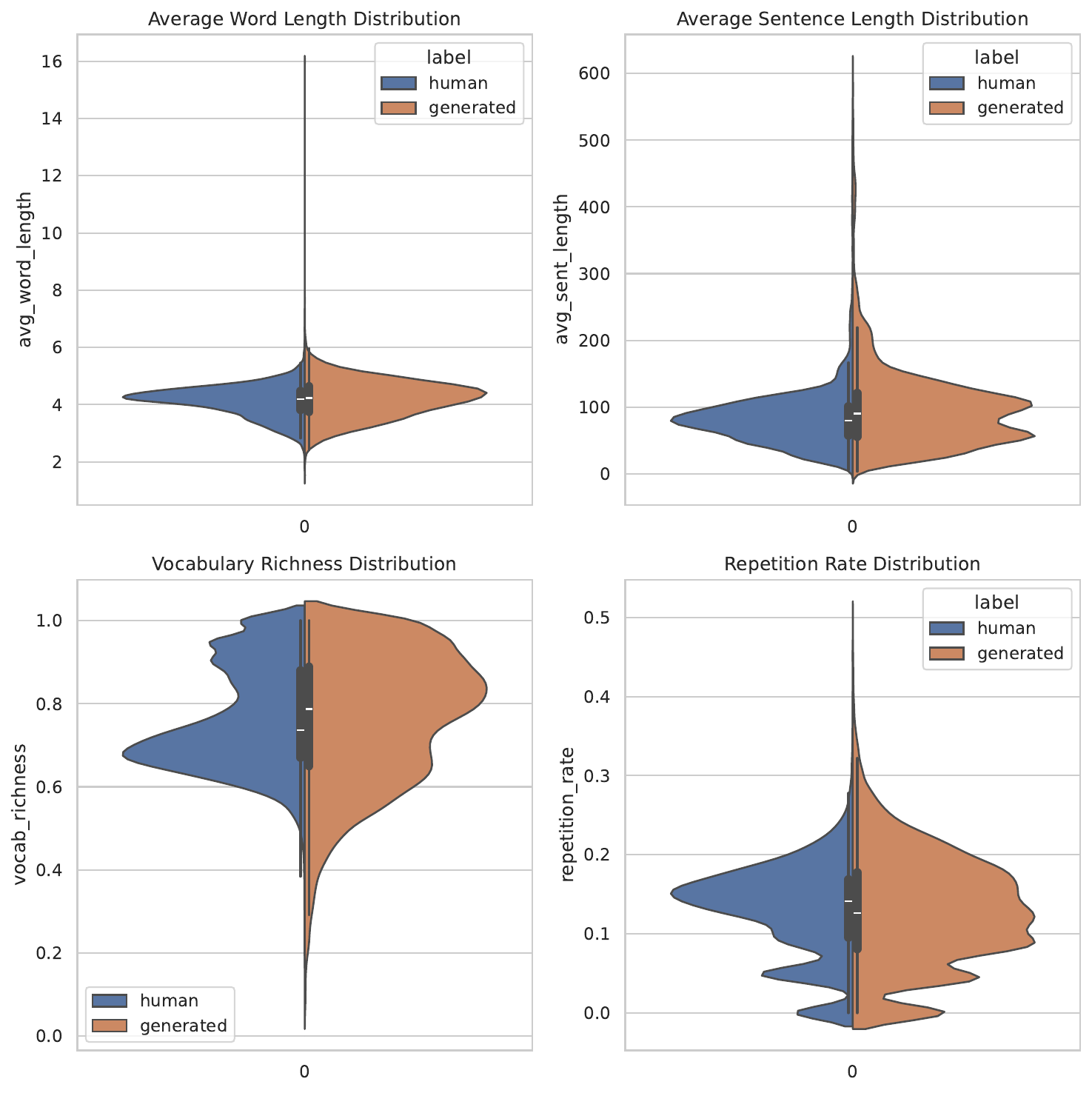}
         \caption{Spanish}
         \label{fig:EDA_subtask1_es}
     \end{subfigure}
        \caption{Subtask 1 feature engineered dataset statistics.}
        \label{fig:EDA_subtask1}
\end{figure*}

Figure \ref{fig:EDA_subtask1} presents a comparative visualization of feature-engineered dataset statistics for Subtask 1, encompassing both English and Spanish languages. The distribution patterns across the datasets for each language are delineated by average word and sentence length, alongside vocabulary richness and repetition rate. Notably, the visualizations elucidate the differences between human-generated and machine-generated text, with the human-generated text typically showcasing greater variability in sentence length and vocabulary richness.

\begin{figure*}[!h]
     \centering
     \begin{subfigure}[b]{0.45\textwidth}
         \centering
         \includegraphics[width=\textwidth]{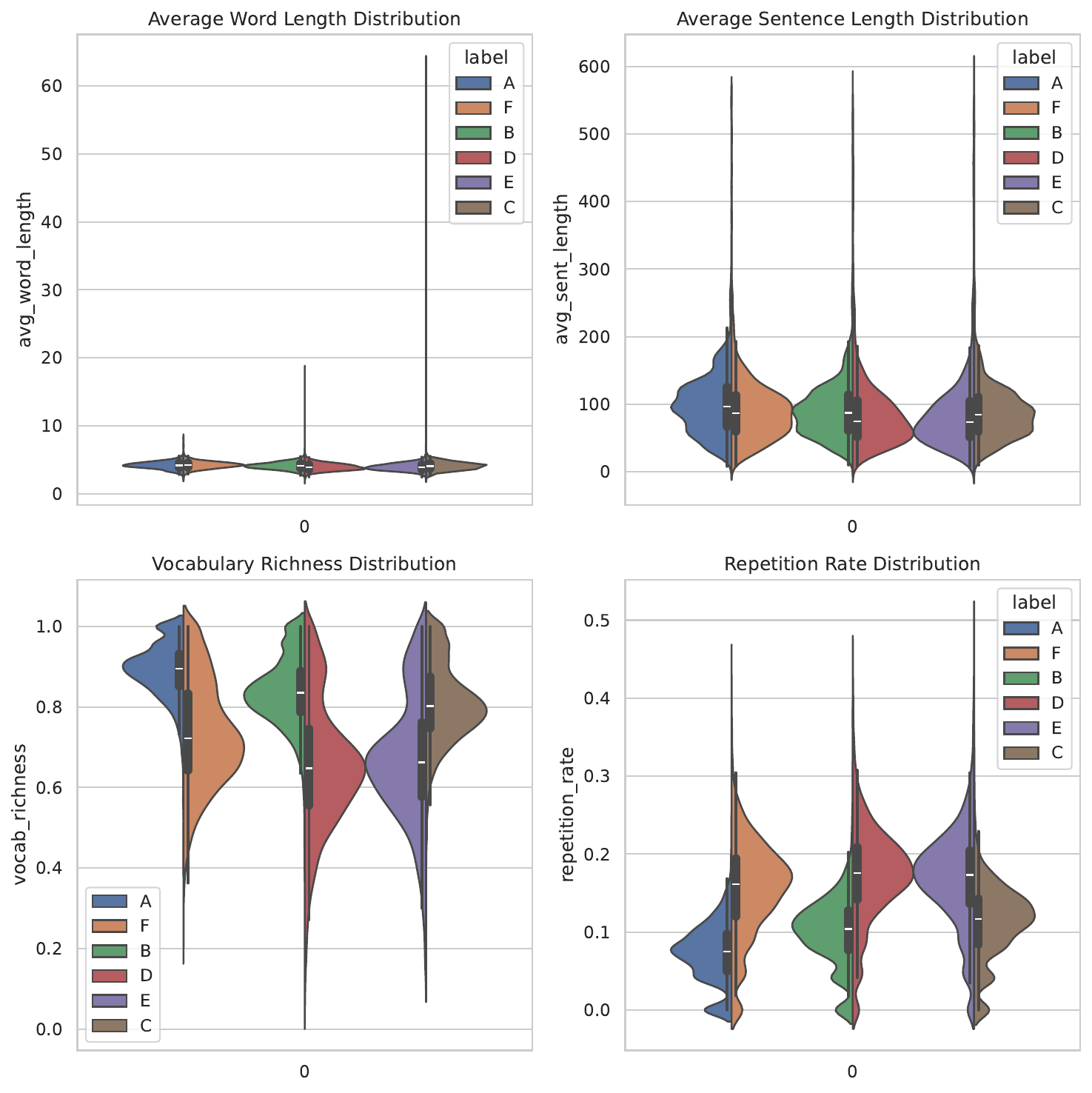}
         \caption{English}
         \label{fig:EDA_subtask2_en}
     \end{subfigure}
     \begin{subfigure}[b]{0.45\textwidth}
         \centering
         \includegraphics[width=\textwidth]{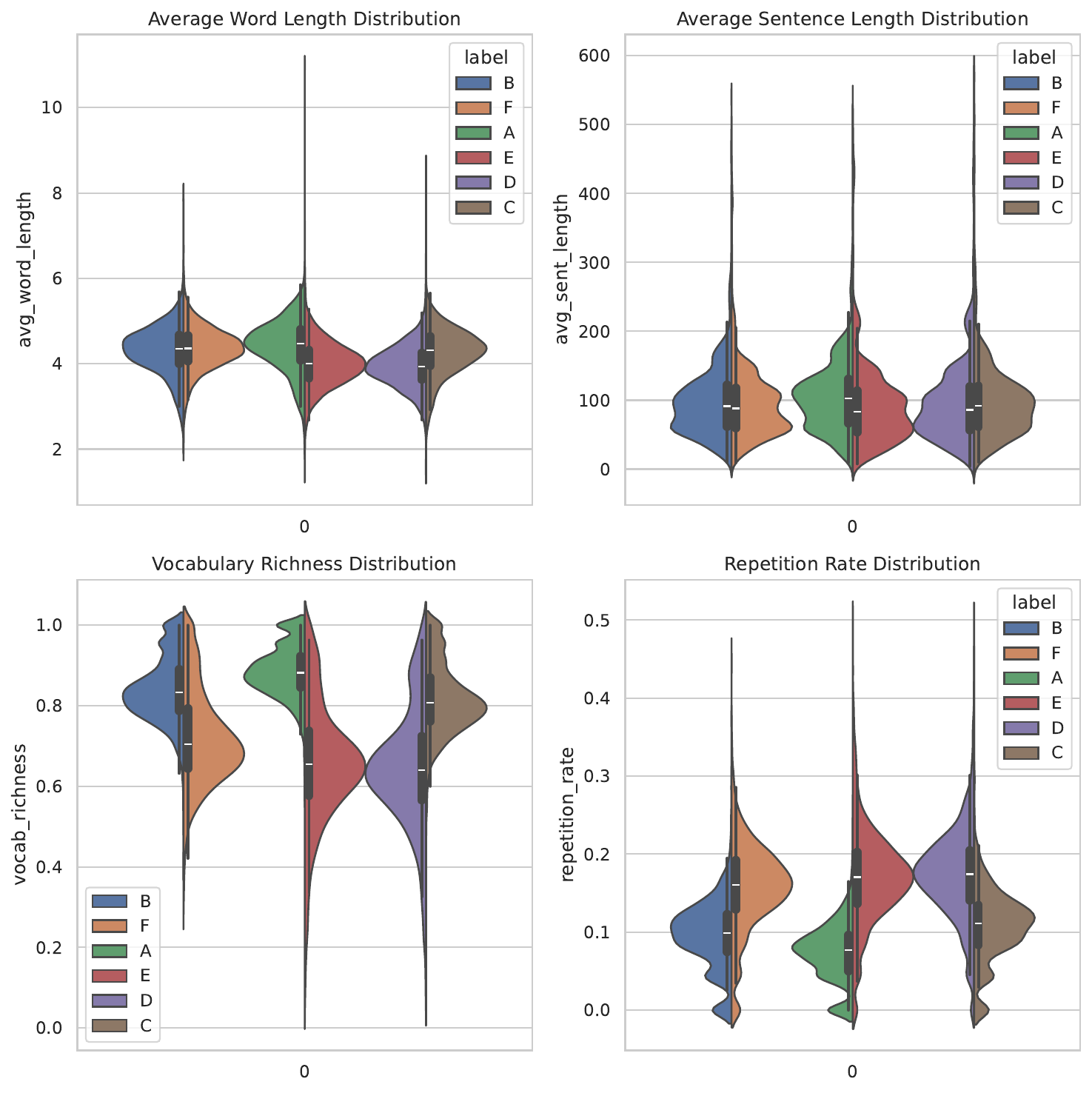}
         \caption{Spanish}
         \label{fig:EDA_subtask2_es}
     \end{subfigure}
        \caption{Subtask 2 feature engineered dataset statistics.}
        \label{fig:EDA_subtask2}
\end{figure*}

Figure \ref{fig:EDA_subtask2} offers a detailed feature comparison for Subtask 2, showcasing statistical analyses of engineered datasets in both English and Spanish. This figure provides insights into the average word and sentence length distributions, as well as vocabulary richness and repetition rate across different labels, significantly expanding upon the foundational comparisons of Subtask 1.

\clearpage
\section{Feature Engineered Dataset Samples}
We present samples from our feature-engineered dataset, which has been specifically curated to facilitate the analysis of textual features that may distinguish between human-generated and machine-generated text. The dataset consists of text snippets, each labeled as either 'human' or 'generated', representing the origin of the text. The features engineered for this analysis include Average Word Length (AWL), Average Sentence Length (ASL), Vocabulary Richness (VR), and Repetition Rate (RR).

Tables \ref{tab:subtask1_en} and \ref{tab:subtask1_es} display subsets of our dataset, illustrating the distribution of these features across texts labeled as 'human' or 'generated'. These samples exhibit the variability within and between categories, forming the basis for subsequent analysis aiming to identify patterns and markers indicative of the text's origin. The engineered features are expected to contribute to the development of models capable of differentiating between human and machine-generated text.

\begin{table*}[hb]
\centering
\resizebox{1\linewidth}{!}{
\begin{tabular}{p{0.55\textwidth}ccccc}
\toprule
\textbf{Text} & \textbf{Label} & \textbf{AWL} & \textbf{ASL} & \textbf{VR} & \textbf{RR} \\
\midrule
you need to stop the engine and wait until it stops. This is how I would do it: // Check if its safe & generated & 3.120 & 49.500 & 0.960 & 0.040 \\
I have not been tweeting a lot lately, but I did in November, and it was a really good month. I also & generated & 3.160 & 49.500 & 0.840 & 0.120 \\
I pass my exam and really thankgod for that but idk where will I go for shsmy result is ah & human & 3.550 & 90.000 & 0.900 & 0.100 \\
@PierreJoye i have a server already, thanks for the offer the problem is time, as always :p (ill be done & human & 3.400 & 104.000 & 0.920 & 0.080 \\
Crying because I have to cry for you?. No. No, no, no. Itll be all right. I & generated & 2.458 & 14.200 & 0.708 & 0.208 \\
\bottomrule
\end{tabular}}
\caption{English feature engineered dataset on Subtask 1.}
\label{tab:subtask1_en}
\end{table*}

\begin{table*}[!hbp]
\centering
\resizebox{1\linewidth}{!}{
\begin{tabular}{p{0.55\textwidth}ccccc}
\toprule
\textbf{Text} & \textbf{Label} & \textbf{AWL} & \textbf{ASL} & \textbf{VR} & \textbf{RR} \\
\midrule
Mam, por qu no me despertaste? Te hable 5 veces, te grite, te prend la luz y te abr	 & human & 2.827 & 41.000 & 0.826 & 0.087 \\
. Artculo 2. Los Estados miembros aplicarn las medidas necesarias para cumplir la presente Directiva a ms tardar el 31 de diciembre de 1981. Artculo 3. Los destinatarios de la presente Directiva sern los Estados miembros. Hecho en Luxemburgo, el 30 de junio de 1981.	 & human & 4.353 & 43.500 & 0.647 & 0.216 \\
Mi memoria es: 5\% de los mdicos tienen una alta vocacin y por lo tanto son buenos profesionales, el resto es prescind	 & generated & 3.840 & 118.000 & 0.960	& 0.040 \\
APROBAR el proyecto de resolucin que se adjunta como Anexo I, por la cual se aprueba la solicitud presentada por el seor Csar Enrique Vega Arvalo (CP N PI:KEY), con domicilio en calle 7 N 3080 Quilicura, comuna de Santiago. Artculo 2. Notifquese y publquese. Dado en La Moneda, a los veintisiete das del mes de diciembre de dos mil diecinueve. Curso de Photoshop CS6 Bsico para & generated & 3.937 & 74.600 & 0.797 & 0.114 \\
De pequeo Dios me dio a elegir entre tener una memoria increble o un pito gigante y no me acuerdo lo que eleg & human & 3.784 & 109.000 & 0.957 & 0.043 \\
\bottomrule
\end{tabular}}
\caption{Spanish feature engineered dataset on Subtask 1.}
\label{tab:subtask1_es}
\end{table*}

\clearpage
\section{Evaluation on Subtask 2}
\begin{figure*}[!htbp]
     \centering
     \begin{subfigure}[b]{0.45\textwidth}
         \centering
         \includegraphics[width=1\textwidth]{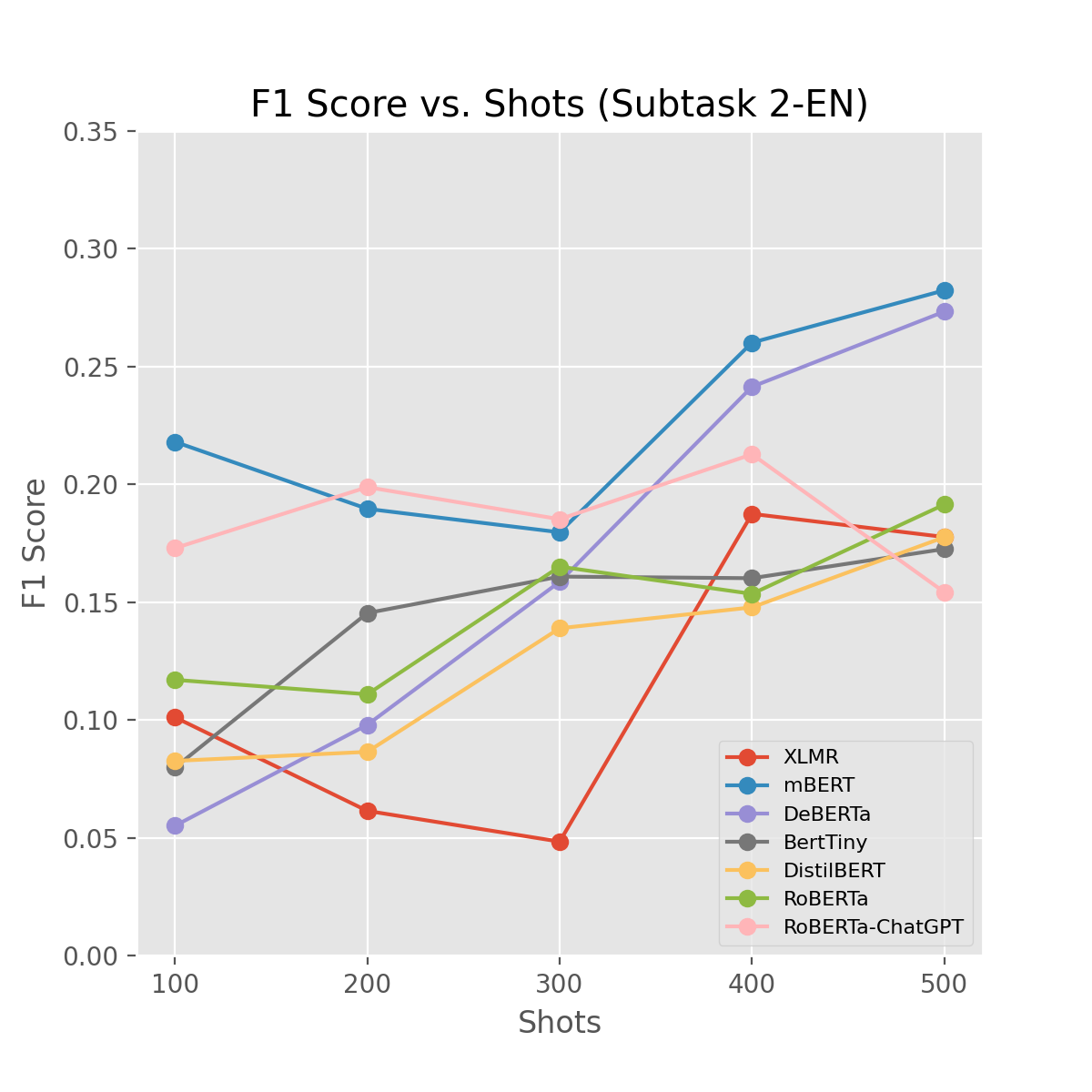}
         \caption{English}
         \label{fig:fewshot_eval_subtask2_en}
     \end{subfigure}
     \begin{subfigure}[b]{0.45\textwidth}
         \centering
         \includegraphics[width=1\textwidth]{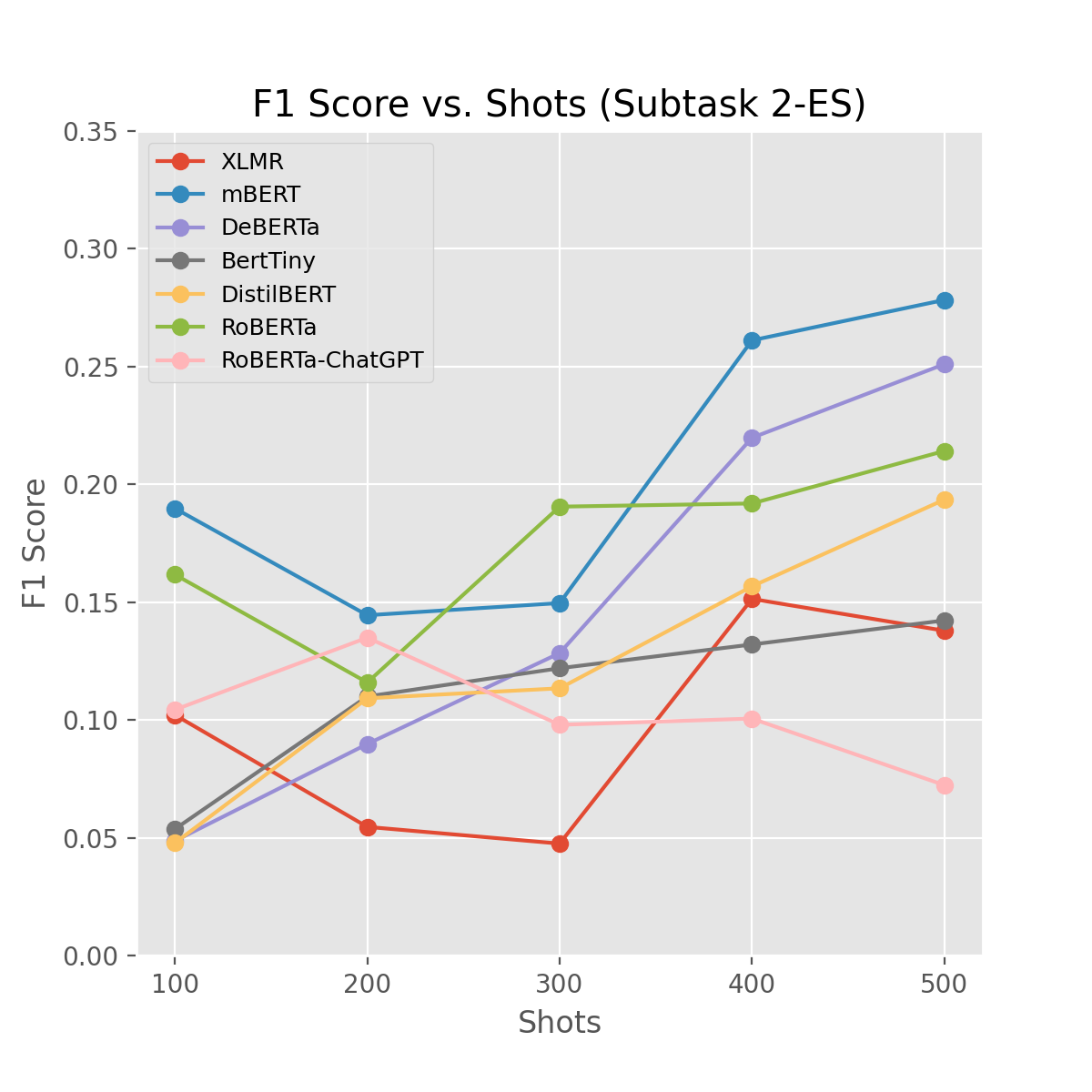}
         \caption{Spanish}
         \label{fig:fewshot_eval_subtask2_es}
     \end{subfigure}
        \caption{Subtask 1 Evaluation on Few-Shot Learning}
        \label{fig:few_shot_subtask2}
\end{figure*}

In Figure \ref{fig:few_shot_subtask2}, we observe the evaluation of few-shot learning performance across various models for Subtask 1 in both English and Spanish, denoted as Subtask 2-EN and Subtask 2-ES respectively. The F1 Score versus the number of shots (examples) is plotted, providing a clear illustration of how model performance scales with the amount of provided training data. Notable trends include the progressive improvement of models like RoBERTa and its variant RoBERTa-ChatGPT with increasing data, as well as the comparatively high performance of XLM-R in both languages.

\end{document}